\documentclass[fleqn,10pt]{olplainarticle}
\usepackage{palatino}
\usepackage[font=small,labelfont=bf,
   justification=justified,
   format=plain]{caption} % 'format=plain' avoids hanging indentation
\usepackage{mathpazo}
\usepackage{bm}
\usepackage{tabularx}
\usepackage{subcaption}
\usepackage{float}
\usepackage[hidelinks]{hyperref}
\usepackage{cleveref}

\title{FALCON: Few-Shot Adversarial Learning for Cross-Domain Medical Image Segmentation}

\author[1]{Abdur R. Fayjie\thanks{Corresponding Author: fayjie92@gmail.com}}
\author[2]{Pankhi Kashyap}
\author[3]{Jutika Borah}
\author[1]{Patrick Vandewalle}
\affil[1]{KU Leuven, Leuven, 3000, Belgium.}
\affil[2]{IIT-Bombay, Mumbai, 400076, India.}
\affil[3]{Tezpur University, Tezpur, 784028, India.}


\begin{abstract}
Precise delineation of anatomical and pathological structures within 3D medical volumes is crucial for accurate diagnosis, effective surgical planning, and longitudinal disease monitoring. Despite advancements in AI, clinically viable segmentation is often hindered by the scarcity of 3D annotations, patient-specific variability, data privacy concerns, and substantial computational overhead. In this work, we propose FALCON, a cross-domain few-shot segmentation framework that achieves high-precision 3D volume segmentation by processing data as 2D slices. The framework is first meta-trained on natural images to \emph{learn-to-learn} generalizable segmentation priors, then transferred to the medical domain via adversarial fine-tuning and boundary-aware learning. Task-aware inference, conditioned on support cues, allows FALCON to adapt dynamically to patient-specific anatomical variations across slices. Experiments on four benchmarks demonstrate that FALCON consistently achieves the lowest Hausdorff Distance scores, indicating superior boundary accuracy while maintaining a Dice Similarity Coefficient comparable to the state-of-the-art models. Notably, these results are achieved with significantly less labeled data, no data augmentation, and substantially lower computational overhead.

\end{abstract}

\begin{document}

\flushbottom
\maketitle
\thispagestyle{empty}

\section{Introduction}

Accurate segmentation of anatomical structures, such as the liver, kidney, heart, and pathological regions like brain tumors in MRI, is critical for diagnosis, treatment planning, and monitoring disease progression, enabling clinicians to assess patient conditions comprehensively and make informed decisions. This task is typically performed manually by radiologists or clinicians, rendering it labor-intensive, time-consuming, and subject to variability. To improve efficiency and consistency, automated segmentation methods based on AI have gained significant interest. 

Artificial Intelligence (AI) with Deep Neural Networks (DNNs), particularly those employing transformer architectures, has shown remarkable progress in general image analysis. However, applying these models directly to medical imaging faces several challenges: These models require substantial computational resources for both training and inference, and their training typically depends on access to large-scale annotations. Particularly for 3D volumes, the manual creation of masks by clinical experts is prohibitively expensive and time-consuming. Generative models that create synthetic data offer a promising solution to data and annotation scarcity, yet their clinical adoption is hindered by the need for rigorous validation and regulatory compliance~\citep{fda1, fda2}. Conventional data augmentation techniques, including rotations, scaling, and intensity adjustments, are widely used but may introduce unrealistic variations that fail to capture clinically relevant features accurately, potentially undermining model reliability in practice~\citep{elgendi2021geometric, pattilachan2022critical, madani2018fast, tirindelli2021ultrasound}. Furthermore, an accurate boundary is crucial in medical image segmentation, as small localization errors can have significant clinical consequences, such as inaccurate tumor measurements leading to surgical catastrophe. Commonly used loss functions, including cross-entropy and Dice loss, treat all pixels uniformly and often do not sufficiently emphasize `boundary' regions, limiting segmentation accuracy at edges~\citep{kervadec2021}.

\begin{figure}[!ht]
    \centering
    \medskip
    \includegraphics[width=1\linewidth]{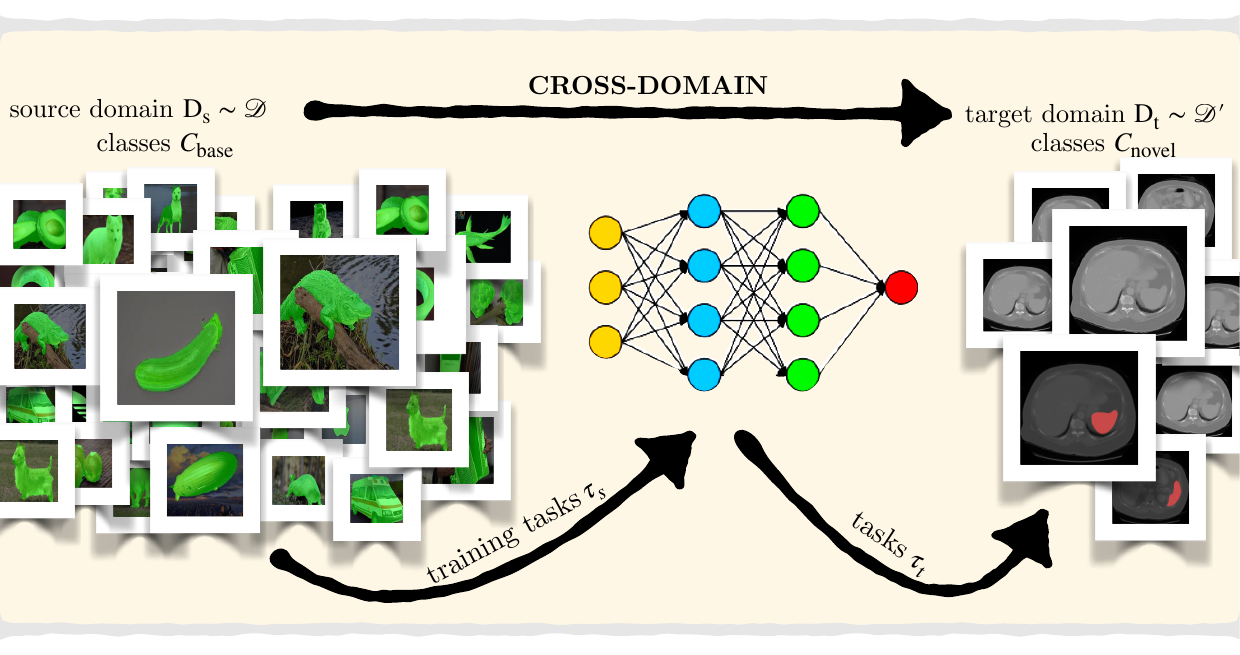}
    \caption{\textbf{Problem Formulation of Cross-Domain Few-Shot Segmentation (CDFSS).} A model is trained on source tasks $\tau_s$, involving base classes $C_\text{base}$ from a source dataset $\mathrm{D}_s \sim \mathcal{D}$ (e.g., natural images). The objective is to generalize to target tasks $\tau_t$ involving previously unseen classes $C_\text{novel}$ from a distinct target dataset $\mathrm{D}_t \sim \mathcal{D}'$ (e.g., medical imaging). The underlying distributions for the source and the target dataset are denoted by $\mathcal{D}$ and $\mathcal{D}'$. This mimics human cognitive processes where medical trainees acquire broad foundational knowledge over time and later adapt it to specialize as clinicians. Unlike the source label-rich source domain, the target domain is characterized by limited data and scarce annotations.}
    \label{fig:falcon_teaser}
\end{figure}

Driven by the need for locally privacy-preserving and resource-efficient medical AI, this paper proposes that unlabeled slices from a 3D volume for a single patient can provide the necessary context for high-accuracy segmentation. We hypothesize that a task-aware inference mechanism enables lightweight models to achieve comparable performance to state-of-the-art (SOTA) methods by leveraging the inherent structural consistency of these unlabeled slices. Consequently, we present a novel framework for Cross-Domain Few-Shot Segmentation (CDFSS) with unlabeled Support. Our formulation relaxes traditional Few-Shot Learning (FSL) requirements to leverage the structural consistency inherent in 3D medical volumes, enabling the model to \emph{learn-to-learn} generalizable priors from natural images and to adapt to medical imaging domains using patient-specific anatomical, textural, and intensity context. Our framework, FALCON (\underline{F}ew-Shot \underline{A}dversarial \underline{L}earning for \underline{C}ross-D\underline{o}main Medical Image Segmentatio\underline{n}), integrates task-based FSL with conventional fine-tuning. This integration is motivated by evidence that fine-tuning enhances performance in FSL tasks~\citep{nakamura2019revisiting, shen2021partial, Wang2023RevisitFS, guo2020cdfsl}. It introduces three key innovations:

\begin{itemize}
    
    \item \emph{Unlabeled Support Integration:} We employ a Relation Module ($RM(\cdot))$ within the network bottleneck to compute affinities between query features and unlabeled support features, effectively treating the support set as a `visual prompt' for patient-specific adaptation.
    
    \item \emph{Boundary-Aware Adversarial Fine-tuning (BAAF):} To ensure high geometric precision, we move beyond standard region-based losses (such as Dice) by incorporating a differentiable Hausdorff Distance (HD) loss. Furthermore, we employ an adversarial training strategy during fine-tuning; a discriminator ensures that predicted masks on unlabeled slices remain anatomically plausible and distributionally consistent with the ground-truth masks.
    
    \item \emph{Task-Aware Inference:} Our approach enables efficient test-time inference utilizing unlabeled support. The model segments an entire 3D volume by conditioning predictions on a few unlabeled slices from the same scan, achieving precise boundary delineation without requiring additional gradient updates.

\end{itemize}

\noindent This lightweight framework is designed for privacy-preserving local deployment, reducing reliance on large-scale annotations and cloud-based AI services.

\section{Related Work}  
\label{sec:related_work_falcon}

\subsection{Cross-Domain Few-Shot Learning}
\label{subsec:falcon_cdfsl_RW}

In many practical applications, the typical assumptions of FSL are violated, as the data distributions between training and test domains may differ significantly. Cross-Domain Few-Shot Learning (CDFSL) addresses this issue by explicitly modeling the base and novel classes as belonging to two distinct data distributions. ~\citet{xu2025cdfslsurvey} provide one of the most distinctive definitions of CDFSL—clearly differentiating it from domain adaptation, domain generalization, multi-task learning, and conventional FSL—which we adopt in our problem formulation (see \cref{sec:falcon_problem_formulation}). Several approaches such as data or feature-based augmentation or adaptation~\citep{adler2021crossdomainfewshotlearningrepresentation, zhao_dual_adaptive, hu2022advcdfsl, Tseng2020Cross-Domain}, task synthesis~\citep{ijcai2021p149}, knowledge distillation~\citep{islam2021dynamic, phoo2021STARTUP}, and regularization~\citep{heidari2024adaptive, cao_semantic_consistency} have been proposed. Our work is particularly motivated by the findings of \citet{guo2020cdfsl}, who demonstrated that fine-tuning outperforms conventional FSL methods on their CDFSL benchmark, which spans a spectrum of datasets ranging from near-domain to distant-domain settings. Additionally, our work closely aligns with \citet{yao2021cdfslunlabelled}, who leverages unlabeled data via self-supervised learning to bridge the gap between source and target domains. However, our approach differs in that we utilize unlabeled data as support examples during the fine-tuning phase. Particularly in medical imaging, recent efforts include FAMNet~\citep{bo2024famnet}, a frequency-matching network proposed to address both intra-domain and inter-domain differences by integrating frequency features to handle shifts between CT and MRI data. \citet{Gong2023-sn} utilize meta-learning via a pseudo-Siamese network to learn from extracted contour features and features from the original images. Their method was evaluated on the benchmark proposed by~\citet{guo2020cdfsl}, considering miniImageNet~\citep{Vinyals2016MN} as the source domain and EuroSAT~\citep{helber2019eurosat} and ChestX~\citep{Wang2019ChestX} as the target domains.

\subsection{Boundary Segmentation} 
\label{sec:falcon_boundary_seg_RW}

The historical quest for precise boundary detection has relied on deformable~\citep{Kass1988, Caselles1997} and atlas-based~\citep{marroquin2002brainmri, park2003atlas} models, which utilize edge information or image registration techniques to minimize global distance-based loss functions. \citet{schmidt2012hd} introduced Hausdorff Distance (HD)-based priors, employing inter-segment constraints to tackle complex multi-surface medical image segmentation tasks. Despite their effectiveness, these priors, formulated as a constrained optimization problem, do not guarantee global optimality due to their reliance on a limited set of feasible solutions. \citet{Karimi2020} pioneered a novel differentiable Hausdorff loss, utilizing distance transforms to enable the direct minimization of the HD via neural networks, which serves as the primary loss function in this work. Building upon this, \citet{celaya2024generalizedsurfacelossreducing} introduced a weighted normalized boundary loss to alleviate class imbalance issues in medical imaging. In contrast to HD-based losses, \citet{kervadec2021} proposed a boundary loss with an unbounded range, spanning from negative to positive infinity, potentially overshadowing the influence of the Dice loss, which is confined to the range [0, 1]. While many other studies emphasize boundary segmentation via architectural innovations~\citep{WANG2022102395} or evaluation metrics~\citep{yin_focal_dice_23, ZAMAN2023107324}, we restrict our discussion to loss-function-based approaches, as they form the methodological foundation of our work.

\subsection{Adversarial Learning} 
\label{subsec:adv_learning_falcon_RW}
Adversarial learning has been a powerful method for training DNNs under labeled data scarcity, particularly in FSL. In this context, as a regularizer, it helps reduce overfitting. Simply put, adversarial learning introduces a max-min optimization problem, where a model (the generator) competes with a discriminator, encouraging the model to learn domain-invariant and more robust representations, thereby enhancing its overall generalization capability. \citet{ganin2016dann} proposed the Domain-Adversarial Neural Network (DANN), a foundational work in adversarial learning, where the gradient reversal layer enforces feature invariance across domains. In medical imaging, \citet{zhang2017biomedadv} proposed leveraging unlabeled data alongside labeled data for biomedical image segmentation, a key motivation for the present study. However, our work introduces adversarial learning as a regularizer during the fine-tuning phase. ~\citet{Chen2020MRIadv} apply adversarial learning in a slightly different context: generating plausible and realistic signal corruptions to model common artifacts in MRI imaging, such as bias fields. In the context of FSL for medical imaging, \citet{Mondal2018Fewshot3M} and \citet{CHEN2022112} employed adversarial learning with a U-Net architecture for 3D and 2D segmentation, respectively. PG-Net~\citep{awudong2024adv} proposes training DNNs without annotations via two subnetworks: P-Net, a prototype-based segmentation network that extracts multi-scale features and local spatial information to produce segmentation maps, and G-Net, a discriminator equipped with an attention mechanism that distills relational knowledge between support and query images. G-Net contributes to P-Net’s ability to generate query segmentation masks with distributions more closely aligned to the support set.

\section{Problem Formulation}
\label{sec:falcon_problem_formulation}

We consider a CDFSS problem, where a model must learn segmentation priors from abundant labeled data in a source domain $\mathrm{D}_s$ (natural images) to generalize to novel structures in a target domain $\mathrm{D}_t$ (medical images) provided minimal supervision. This setup follows the CDFSL framework formalized by~\citet{xu2025cdfslsurvey}: (i) $\mathrm{D}_s$ and $\mathrm{D}_t$ are sampled from from distinct underlying distributions $\mathcal{D}$ and $\mathcal{D'}$ ($\mathcal{D} \neq \mathcal{D'}$), respectively; (ii) The base classes $C_\text{base}$ in $\mathrm{D}_s$ share no overlap with the novel classes $C_\text{novel}$ in $\mathrm{D}_t$; (iii) $\mathrm{D}_t$ contains significantly fewer samples than $\mathrm{D}_s$; and (iv) $\mathrm{D}_t$ contains annotations for only a limited fraction of the available target samples. 

\noindent The source dataset $\mathrm{D}_s$, containing $n_s$ samples, is defined as:

\begin{equation}
    \mathrm{D}_s = \{\bm{x}_i, \bm{y}_i\}_{i=1}^{n_s},
\end{equation}

\noindent where $\bm{x} \in \mathbb{R}^{H \times W \times 3}$ represents a 3-channel RGB image of height $H$ and width $W$, and $\bm{y} \in [0,1]^{H \times W}$ denotes its corresponding ground-truth segmentation mask. 

The target domain consists of medical images (CT or MRI slices) from $\Pi$ patients. For each patient 
$\pi \in \{1, \dots, \Pi\}$, we define:

\begin{equation}
\mathrm{D}_t^{(\pi)} = 
\underbrace{\{\bm{x}_j\}_{j=1}^{n_u}}_{\text{unlabeled}}
\cup 
\underbrace{\{(\bm{x}_i, \bm{y}_i)\}_{i=1}^{n_l}}_{\text{labeled}},
\label{eq:patient_data_split}
\end{equation}
with $n_u >> n_l$.

The target domain problem is formulated as a binary segmentation, represented as a collection of $1$-way $K$-shot tasks. Each task, $\tau_t$ is sampled from an underlying task distribution $\mathcal{T}$, which in practice, is instantiated through a finite collections of tasks constructed from $\mathrm{D}_t \sim \mathcal{D}'$. $\tau_t$ is composed as a pair of two distinct sets: the Support Set $\mathrm{S}$ and the Query Set $\mathrm{Q}$. In contrast to conventional FSL, where the support set consists of $K$ labeled samples, our setting uses a support set of $K$ unlabeled samples, which the model uses for adaptation with minimal supervision by calculating relation-based prototypes (see \cref{sec:falcon_framework}). The query set consists of test samples during inference; however, during fine-tuning, it is composed of labeled examples drawn from the small annotated subset in \cref{eq:patient_data_split}. Consequently, $\tau_t = (\mathrm{S}, \mathrm{Q})$ is constructed per patient during fine-tuning:

\begin{equation}
    \mathrm{S} = \{\bm{x}_j\}; \ j = 1, \dots, K \qquad \text{and} \qquad
    \mathrm{Q} = \{\bm{x}_q, \bm{y}_q\}.
    \label{eq:target_task_def}
\end{equation}

%\noindent $\mathrm{S}$ samples $K$ unlabeled slices from the same patient and anatomical structure. Critically, no pixel-wise labels are assumed for $\mathrm{S}$, making this distinct from classical FSL. $\mathrm{Q}$ samples labeled slice(s) for the same patient from the labeled set, used for supervision during fine-tuning.
This formulation is patient-specific: both $\mathrm{S}$ and $\mathrm{Q}$ are drawn from the same patient $\pi$, ensuring anatomical and acquisition consistency, an assumption validated in clinical practice where 3D volumes yield multiple co-registered 2D slices.

The goal is to learn a segmentation model $f_\theta:\mathbb{R}^{H \times W \times 3} \rightarrow [0,1]^{H \times W}$ such that: (i) It is first trained on (metric-based meta-training) $\mathrm{D}_s$ to \emph{learn-to-learn} general segmentation priors. (ii) It is then adapted to $\mathrm{D}_t$ by leveraging both labeled queries and unlabeled support from the same patient. This phase employs Boundary-Aware Adversarial Fine-tuning (BAAF), and (iii) At test-time, given a new patient $\pi'$ not seen during fine-tuning, the model segments query slices conditioned on unlabeled support for $\pi'$, without any gradient updates via task-aware inference.

\section{Proposed Framework}
\label{sec:falcon_framework}

\begin{figure}[!ht]
    \centering
    \medskip
    \includegraphics[width=1\linewidth]{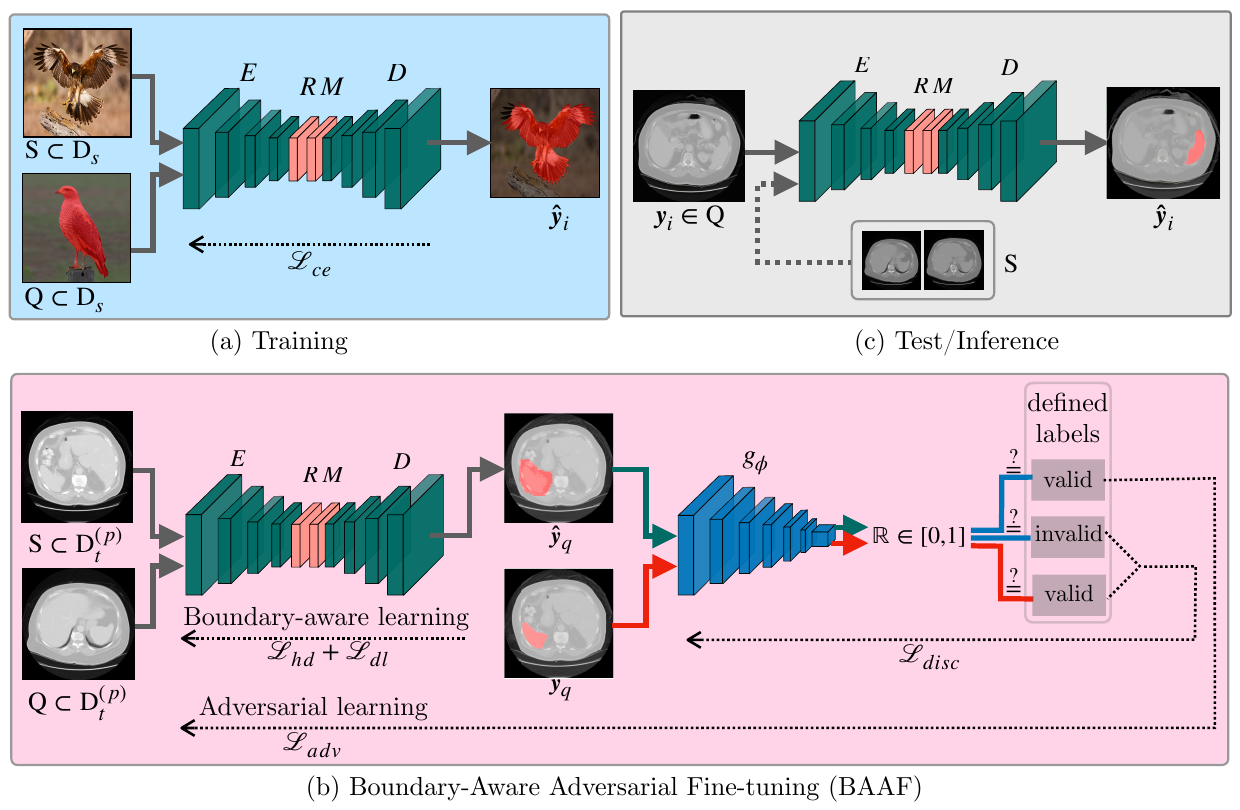}
    \caption[Overview of FALCON]{Overview of FALCON for precise boundary segmentation in medical imaging: (a) \textbf{Training} phase using abundant annotated natural images from the source domain $\mathrm{D}_{s}$, enabling the model to \emph{learn to learn} segmentation knowledge. (b) In the next phase,  \textbf{Boundary-Aware Adversarial Fine-tuning (BAAF)} adapts the model to the target medical domain $\mathrm{D}_{t}$ by leveraging a small annotated subset of slices along with a large collection of unlabeled slices as the support set for a patient $\pi$. (c) \textbf{Test/Inference} segments entire slices for a patient previously unseen during fine-tuning, while leveraging selective slices as an unlabeled support set, leading to task-aware inference. The segmentation network $f_\theta$ consists of three key components: an encoder ($E$), a relation module ($RM$), and a decoder ($D$).}
    \label{fig:falcon_framework}
\end{figure}

We propose the FALCON framework, designed for precise boundary delineation of previously unseen medical structures under limited supervision. \Cref{fig:falcon_framework} illustrates an overview of the framework and its key operational phases: training, fine-tuning, and test (and inference). The model $f_\theta(\cdot)$ is based on a U-Net architecture comprising three key components: an encoder $E(\cdot)$, a relation module $RM(\cdot)$, and a decoder $D(\cdot)$. Let $L$ denote the total number of downsampling/upsampling layers in the U-Net architecture.

\paragraph{Encoder.} The encoder $E(\cdot)$, instantiated with EfficientNet-B0~\citep{pmlr-v97-tan19a}, extracts hierarchical feature maps from an input image. For an input image $\bm{x}\in \mathrm{Q} \cup \mathrm{S}$, $E_l(\bm{x})$ represents the feature map extracted by the $l$\textsuperscript{th} layer. The bottleneck feature map, at the deepest level (layer $L$), is denoted as $E_L(\bm{x})$.

\paragraph{Relation Module.} Inspired by Relation Networks for few-shot classification~\citep{Sung2018RN}, our framework adapts this approach to the segmentation task. Specifically, it employs a relation module into the bottleneck of a U-Net architecture which aggregates support features into a single patient-specific prototype. The query representation is then conditioned on this prototype to guide precise segmentation. During fine-tuning and test-time inference, this module enables patient-specific adaptation by leveraging \emph{unlabeled} support slices from the same patient as contextual priors, without requiring pixel-wise annotations.

Given a query image $\bm{x}_{q}$ and a support set $\mathrm{S} =\{\bm{x_j}\}_{j=1}^K$ of $K$ unlabeled 2D slices from the same patient, the encoder $E(\cdot)$ extracts bottleneck feature maps:

\begin{equation}
    F_\mathrm{Q} = E_L(\bm{x}_q) \in \mathbb{R}^{m \times H' \times W'}, \qquad
    F_{\mathrm{S}_j} = E_L(\bm{x}_j) \in \mathbb{R}^{m \times H' \times W'},
    \label{eq:sup_query_finetune}
\end{equation}

\noindent where $j = 1, \dots, K$. To form relational representations, the support features are first aggregated into a single patient-specific prototype:

\begin{equation}
    F_{\mathrm{S}}^{\text{proto}} = \sum_{j=1}^K F_{\mathrm{S}_j} \in \mathbb{R}^{m \times H' \times W'}
    \label{eq:patient_specific_prototype}
\end{equation}

\noindent This prototype is then concatenated with the query feature map along the channel dimension to produce the final relation representation:

\begin{equation}
F_{\mathrm{rel}} = \text{concat}[F_{\mathrm{Q}}; F_{\mathrm{S}}^{\text{proto}}] \quad \in \mathbb{R}^{2m \times H' \times W'},
\end{equation}
where $[;]$ denotes channel-wise concatenation.
The resulting tensor $F_{\mathrm{rel}}$ serves as the input to the deepest layer of the decoder, effectively conditioning the segmentation of the query on structural and textural cues from the unlabeled support slices. This design ensures that all adaptation to the new patient is label-efficient. 

\paragraph{Decoder (D).} 
The decoder, $D(\cdot)$, instantiated with a U-Net decoder, takes the relation pairs as input in its deepest layer.

\begin{equation}
\begin{aligned}
\text{Input to deepest decoder layer:} \quad & F_{\mathrm{rel}} \in \mathbb{R}^{2m \times H' \times W'}, \\
\text{Output of deepest decoder layer:} \quad & F_{\mathrm{dec}}^L = D_L(F_{\mathrm{rel}}),
\end{aligned}
\end{equation}

\noindent In its subsequent layers, it progressively upsamples the features, integrating information from the encoder via skip connections. For layers $l = L-1, \dots, 1$, the upsampled features from the previous decoder layer are concatenated with the corresponding skip connection from the encoder and then processed by the decoder block.

\begin{equation}
    F^l_{dec} = D_l\Big([U(F^{l+1}_{dec}); E_l(\bm{x}_q)]\Big).
\end{equation}

\noindent Here, $D_l$ denotes the operation of the $l$\textsuperscript{th} decoder block, $\mathrm{U}$ denotes the upsampling operation, and $E_l(\boldsymbol{x}_q)$ represents the feature map from the $l$\textsuperscript{th} encoder layer for the query image $\bm{x}_q$, serving as the skip connection. 

\noindent The final segmentation map $\hat{\bm{y}}_q$ is obtained by applying a $1 \times 1$ convolutional layer and an activation function $\sigma$ (\emph{sigmoid} in our experiments) to the output of the shallowest decoder layer,

\begin{equation}
    \hat{\bm{y}}_q = \sigma (\text{conv}_{1\times1} (F^1_{dec})).
\end{equation}

The rationale behind this architectural choice stems from two key considerations: (i) the inherent challenges posed by real-world pixel-wise annotation scarcity for medical data, and (ii) the necessity for computational efficiency to enable deployment on resource-constrained edge devices.

\subsection{Training} 
\label{subsec:falcon_meta_training}

The training phase mimics the FSL task definition of the subsequent fine-tuning phase by employing an episodic training paradigm. Each episode samples a $1$-way $K$-shot training task, $\tau_{s} = \{\mathrm{S}, \mathrm{Q}\}$ using a class chosen from $C_\text{base}$ classes in the source dataset $\mathrm{D}_{\text{s}}$. The support set, $\mathrm{S}$ then becomes,
\begin{equation} 
\mathrm{S} = \{(\boldsymbol{x}_{i}, \boldsymbol{y}_{i}) \mid i=1, \dots, K\} \subset \mathrm{D}_{s}.
\end{equation}

Note that during this phase, support sets are fully labeled. The query set $\mathrm{Q}$ contains disjoint examples which belong to the same class. In our setting, $\mathrm{D}_{s}$  corresponds to the FSS-1000 dataset~\citep{Li_2020_CVPR}. The segmentation network $f_\theta(\cdot)$ is trained on these tasks by optimizing standard cross-entropy loss to learn transferable priors that support cross-domain medical image segmentation.

\subsection{Boundary-Aware Adversarial Fine-Tuning}
\label{subsec:falcon_meta-fine-tuning}
Following the above training phase, the framework undergoes fine-tuning on patient-specific target tasks $\tau_t = (\mathrm{S}, \mathrm{Q})$, as defined in \cref{eq:target_task_def}. %Note that during this phase, support sets are unlabeled. 
This phase employs BAAF, a dual-optimization strategy that integrates boundary-aware learning to ensure structural anatomical precision coupled with adversarial learning to leverage the unlabeled support set.

\paragraph{Boundary-Aware learning.} To achieve precise boundaries in segmentation, we employ boundary-aware learning using the Hausdorff loss~\citep{Karimi2020}. The optimization of this loss function aims to minimize the discrepancy between the predicted and ground-truth masks at the boundary level, thereby enhancing pixel-level boundary accuracy. It is defined as,

\begin{equation}
\mathcal{L}_{\text{hd}} = 
\frac{1}{|\Omega|} \sum_{x \in \Omega} \left( 
\hat{\boldsymbol{y}}_q(x) \cdot d_{\boldsymbol{y}_q}(x)^a + 
\boldsymbol{y}_q(x) \cdot d_{\hat{\boldsymbol{y}}_q}(x)^a 
\right)
+ \lambda_1 \left( 1 - \frac{2 \displaystyle\sum_{\Omega} (\hat{\boldsymbol{y}}_q \odot \boldsymbol{y}_q)}{\displaystyle\sum_{\Omega} (\hat{\boldsymbol{y}}_q^2 + \boldsymbol{y}_q^2)} \right),
\label{eq:hd_loss}
\end{equation}

\noindent where $\Omega$ denotes all pixel grids on which the image is defined, $\odot$ represents the element-wise Hadamard product, and $d$ refers to the distance maps, computed as unsigned distance to the corresponding object boundaries. The parameter $a$ is a penalty coefficient that controls the degree to which larger errors are penalized. The second part of \cref{eq:hd_loss} represents the Dice loss, weighted by the factor $\lambda_1$. Jointly optimizing the Hausdorff loss with the Dice loss helps achieve stability during training, particularly in the initial stages.

\paragraph{Adversarial learning.} 
Our framework integrates a discriminator network $g_\phi(\cdot)$ (in fine-tuning) that takes a segmentation mask (a spatial probability map in [0, 1]) as input and outputs a scalar probability that the mask is real. This adversarial component mitigates the lack of supervision due to the unlabeled support set. The discriminator is implemented as a 2D-CNN, where each convolutional layer (except the first) is followed by batch normalization, a Leaky ReLU activation with a negative slope of 0.2, and dropout with a rate of 0.25. The final layer applies a sigmoid activation to produce a probability score in [0,1], which is used in the adversarial loss. Given $g_\phi(\cdot)$, adversarial loss is defined as follows:

\begin{equation}
    \mathcal{L}_{adv} = \mathbb{E}_{\bm{x}_q} [- \log(g_\phi(f_\theta(\bm{x}_q))].
\end{equation}

\paragraph{Final objective function.}
The segmentation network $f_\theta$ is trained to minimize:
\begin{equation}
    \mathcal{L}_{seg} = \mathcal{L}_{hd} + \lambda_2\mathcal{L}_{adv},
    \label{eq:segmentation_loss}
\end{equation}
while the discriminator $g_\phi$ is trained to maximize:
\begin{equation}
    \mathcal{L}_{disc} = \mathbb{E}_{\boldsymbol{y}_q} [\log g_\phi (\boldsymbol{y}_q)] + \mathbb{E}_{\boldsymbol{x}_q} [\log (1-g_\phi(f_\theta(\boldsymbol{x}_q)))].
\end{equation}
$\lambda_2$ in \cref{eq:segmentation_loss} is the weight factor for the adversarial loss. The discriminator $g_\phi$ acts as an adaptive regularizer, which encourages the predicted masks to resemble real (ground-truth) segmentation masks in terms of structural plausibility and boundary realism.

\subsection{Task-Aware Test and Inference}
\label{subsec:falcon_test_inference}

At test time, FALCON segments images from a previously unseen patient $\pi \notin \{1, \dots, \Pi\}$ during fine-tuning, operating entirely without pixel-wise labels. The model $f_\theta$, adapted to the target domain via BAAF (see \cref{subsec:falcon_meta-fine-tuning}), now performs single-pass, patient-specific inference by leveraging unlabeled intra-patient context.

For a new patient $\pi'$, we form a patient-specific inference task $\tau_\text{infer}^{(\pi')} = (\mathrm{S}^{(\pi')}, \mathrm{Q}^{(\pi')})$, where the support set $\mathrm{S}^{(\pi')} = \{\bm{x}_j\}_{j=1}^{K}$ comprises $K$ unlabeled 2D slices sampled from the patient's 3D volume, and the query set $\mathrm{Q}^{(\pi')}$ includes all slices to be segmented.

Crucially, no optimization occurs at test time. Instead, for each query slice $\bm{x}_q \in \mathrm{Q}^{(\pi')}$, the segmentation is produced by conditioning the network on the support set $\mathrm{S}^{(\pi')}$, through the relation module, exactly as during fine-tuning. Specifically, the support features are aggregated into a single patient-specific prototype (\cref{eq:patient_specific_prototype}), which is fused with the query representation to guide the decoder. This enables implicit, label-free adaptation: the model leverages anatomical and textural consistency across the patient’s own unlabeled slices to enhance boundary precision, without any gradient updates or test-time training.

For evaluation, ground-truth masks are assumed available for computing metrics (see \cref{subsec:evaluation_protocols}). In clinical deployment, however, FALCON operates end-to-end without any annotations, fulfilling its goal of practical few-shot segmentation under extreme label scarcity.

\section{Data}
\label{sec:falcon_data}

\paragraph{FSS-1000.} 
FSS-1000 is a natural image dataset specifically designed for few-shot segmentation and maintains a balanced class distribution. It comprises five support images per class for 1,000 object classes, each with pixel-wise ground-truth annotation. The dataset spans a wide variety of categories, including small everyday objects, cartoon characters, and logos, thereby promoting more robust and generalizable feature learning for FSL models. In our setting, this dataset serves as the source domain $\mathrm{D}_\text{s}$, employed during the meta-training stage.

The following are the medical imaging datasets, each of which serves as a target domain $\mathrm{D}_{t}$ during the meta fine-tuning stage:

\begin{figure}[p]
    \centering
    \medskip
    \begin{subfigure}{0.81\linewidth}
    \includegraphics[width=1\textwidth]{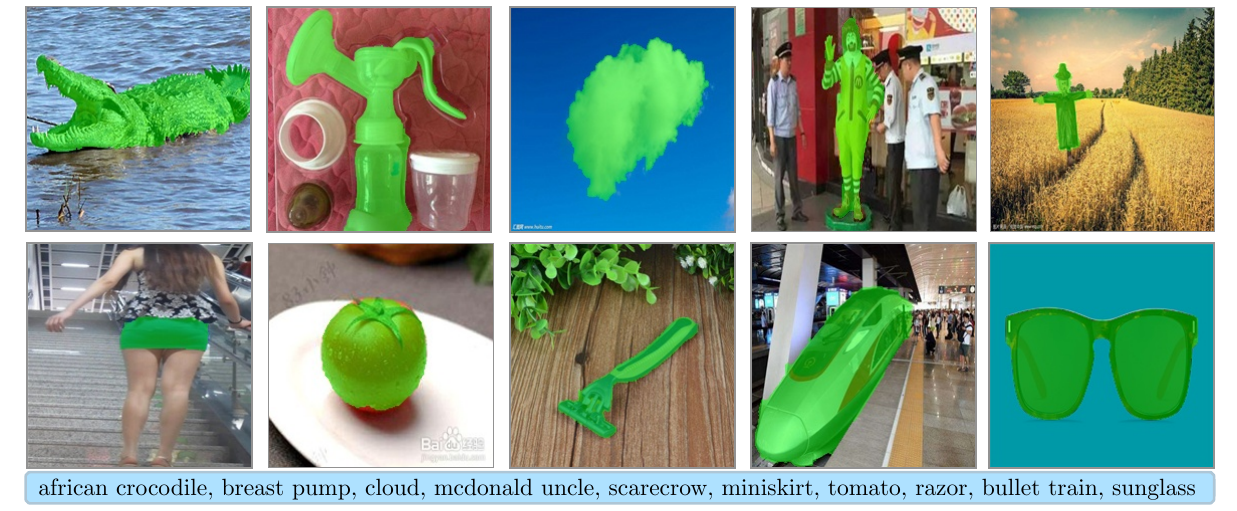}
    \caption{}
    \label{fig2:falcon_0}
    \end{subfigure}
    \vfill
    \begin{subfigure}{0.81\textwidth}
    \includegraphics[width=1\textwidth]{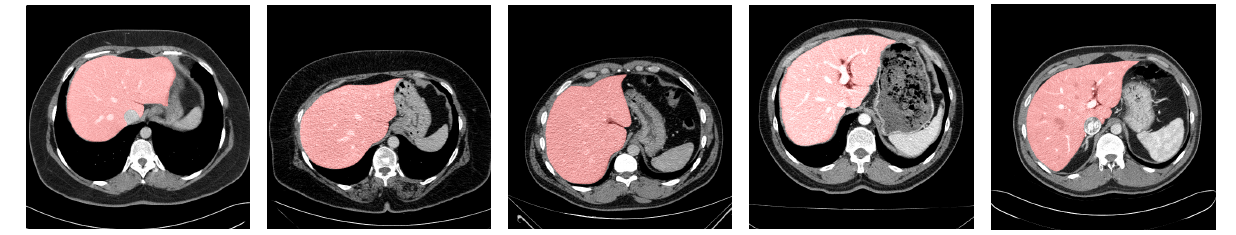}
    \caption{}
    \label{fig2:falcon_a}
    \end{subfigure}
    \vfill
    \begin{subfigure}{0.81\textwidth}
    \includegraphics[width=1\textwidth]{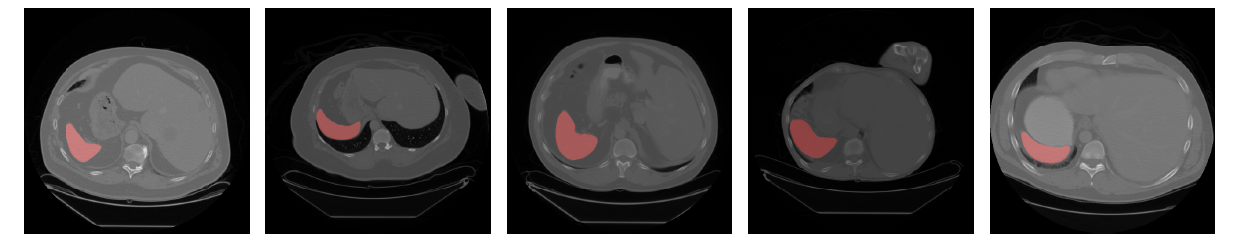}
    \caption{}
    \label{fig2:falcon_b}
    \end{subfigure}
    \vfill
    \begin{subfigure}{0.81\textwidth}
    \includegraphics[width=1\textwidth]{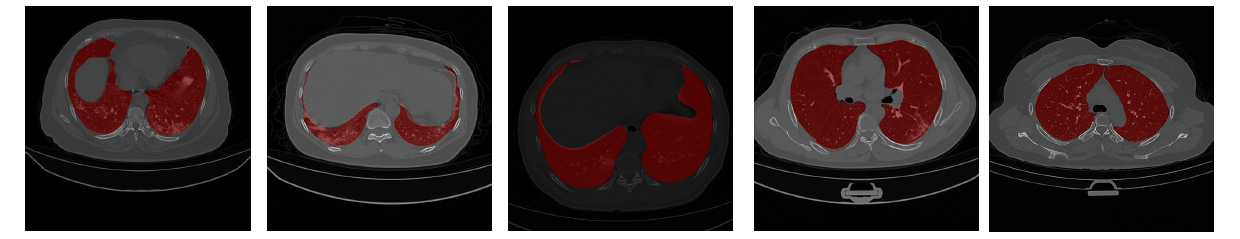}
    \caption{}
    \label{fig2:falcon_c}
    \end{subfigure}
    \vfill
    \begin{subfigure}{0.81\textwidth}
    \includegraphics[width=1\textwidth]{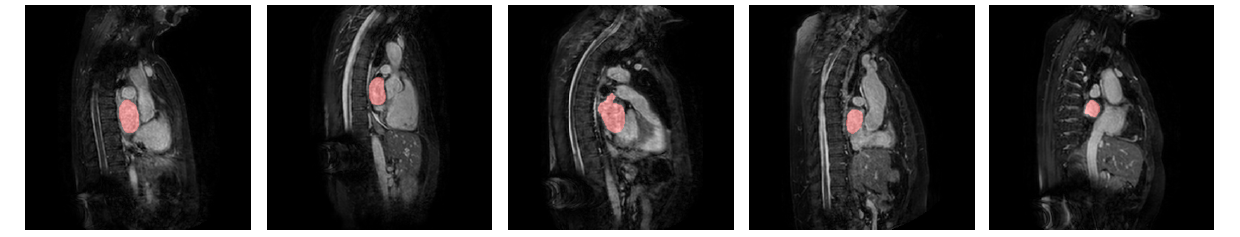}
    \caption{}
    \label{fig2:falcon_d}
    \end{subfigure}
    \caption[Datasets]{Datasets used in our experimental setup: (a) FSS-1000, a natural image dataset illustrated with 10 example classes; and the medical image datasets: (b) CHAOS-CT for liver segmentation, (c) Spleen CT for spleen segmentation, (d) COVID-19 CT for lung infection segmentation, and (e) Cardiac MRI for left atrium segmentation. Segmentation masks are shown in green for FSS-1000 and in red for the medical datasets.}
    \label{fig:falcon_datasets}
\end{figure}

\paragraph{CHAOS-CT.} 
The CHAOS-CT dataset~\citep{kavur2021chaos} contains CT images of 40 potential liver donors with healthy liver (no tumors, lesions, or any other diseases). The images are acquired from the upper abdomen area, 70-80 seconds after contrast agent injection or 50-60 seconds after bolus tracking. Three modalities, a Philips SecuraCT with 16 detectors, a Philips Mx8000 CT with 64 detectors, and a Toshiba AquilionOne with 320 detectors, are used to record data from the subjects in the same orientation and alignment. Each subject's data is represented in 16-bit DICOM images with a resolution of $512\times 512$ pixels, $\mathrm{x}-\mathrm{y}$ spacing of 0.7-0.8mm, and an inter-slice distance of 3 to 3.2 mm. This corresponds to an average of 90 slices per subject, with a minimum of 77 and a maximum of 105 slices. In our setting, the dataset for \emph{liver segmentation} consists of 2,094 slices from 31 patients for training, 172 slices from 4 patients for validation, and 227 slices from 5 patients for testing. Among the training slices, 1,272 out of 2,094 are unlabeled.

\paragraph{Spleen-CT.} 
The Spleen-CT dataset~\citep{simpson2019large} comprises CT scans from 61 patients undergoing chemotherapy treatment for liver metastases at Memorial Sloan Kettering Cancer Center in New York, USA. The CT acquisition and reconstruction follow the following criteria: 120 kVp, 500-1100 ms exposure time, 33-440 mA tube current. Images were reconstructed using a standard convolutional kernel at a thickness varying from 2.5 to 5 mm with a reconstruction diameter range of 360-500 mm. The annotation was performed semi-automatically by segmenting it using the Scout application~\cite{van20073d}. An expert abdominal radiologist manually adjusted the image's contour. In our setting for \emph{spleen segmentation}, the training set comprises 3,000 slices from 50 patients, of which 1,825 slices are unlabeled. The validation set contains 315 slices from 5 patients, and the testing set includes 335 slices from 6 patients.

\paragraph{COVID-19 (CT).} 
The COVID-19 (CT) dataset~\citep{ma2021toward} collects 20 public COVID-19 CT scans from the Corona cases Initiative and Radiopaedia that contain COVID-19 infections. The extent of lung infection ranges from 0.01\% to 59\%. Initial annotations of the left lung, right lung, and infection regions were produced by junior annotators with 1–5 years of experience and subsequently refined by two radiologists with 5–10 years of experience. Finally, all annotations were verified and enhanced by a senior radiologist with over 10 years of expertise in chest radiology. The annotations were manually generated in ITK-SNAP using a slice-by-slice approach on axial images, covering both normal and pathological regions within the whole-lung mask. For \emph{lung segmentation}, we use a dataset comprising 2,175 slices from 14 patients for training (1,313 unlabeled), 150 slices from 2 patients for validation, and 301 slices from 4 patients for testing.

\paragraph{Cardiac-MRI.} 
The Cardiac-MRI dataset~\citep{simpson2019large} contains MRI scans from 30 patients, covering the entire heart during a single cardiac phase, i.e., free breathing with respiratory and ECG gating. Scans were obtained using a 1.5T Achieva scanner (Philips Healthcare, the Netherlands) with resolution $1.25\times1.25\times2.7$ mm\textsuperscript{3}. This dataset was first provided by King’s College London and released publicly as part of the Left Atrial Segmentation Challenge (LASC). Annotations for the left atrial appendage, mitral plane, and portal vein endpoints were initially generated using the automated tool~\cite{ecabert2011} and subsequently refined manually by an expert. For \emph{left atrium segmentation}, the dataset comprises 1,990 training slices from 25 patients (1,190 unlabeled), 105 validation slices from 2 patients, and 285 test slices from 3 patients.

\Cref{fig:falcon_datasets} shows visual samples from the datasets used in our experiments: (a) FSS-1000, a natural image dataset illustrated with 10 example classes; and the medical datasets: (b) CHAOS-CT for liver segmentation, (c) Spleen-CT for spleen segmentation, (d) COVID-19 CT for lung infection segmentation, and (e) Cardiac MRI for left atrium segmentation. Segmentation masks are shown in green for FSS-1000 and in red for the medical datasets.

\begin{figure}[!htbp]
    \centering
    \includegraphics[width=1\linewidth]{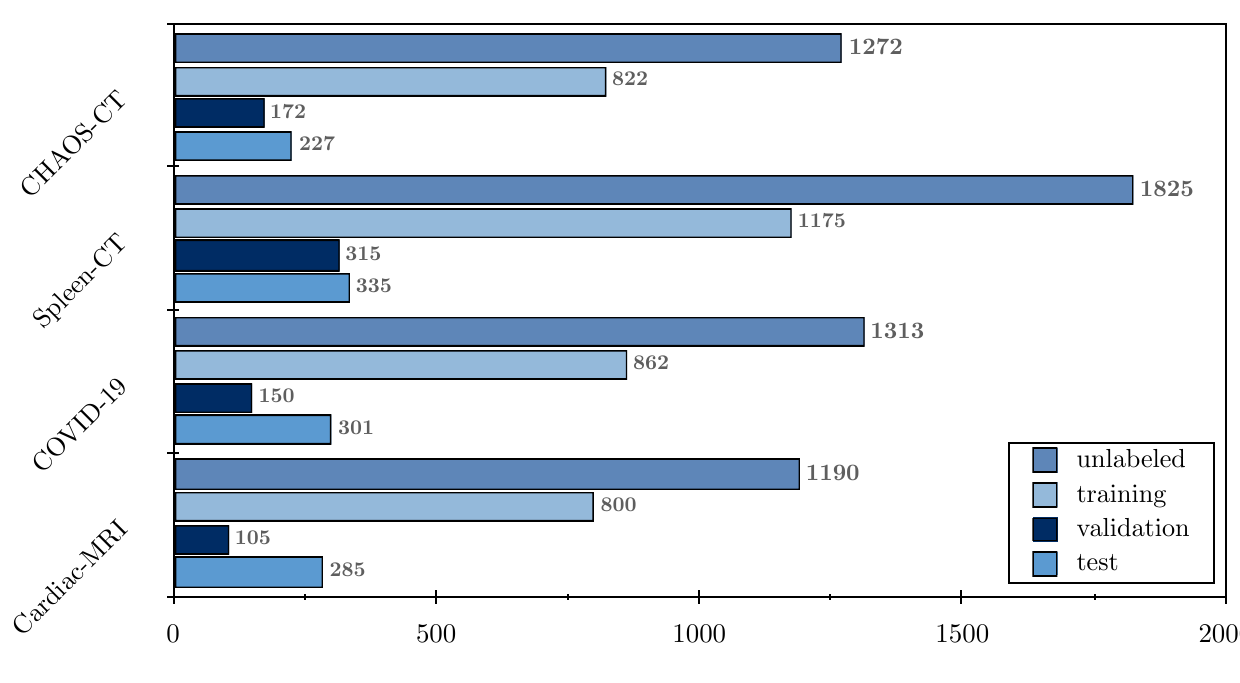}
    \caption[Data distribution, unlabeled and labeled sets]{Distribution of unlabeled and labeled samples across the four medical imaging datasets used in this study. Each horizontal bar represents the number of samples categorized as unlabeled and labeled, further divided into training, validation, and test  sets for the CHAOS-CT, Cardiac-MRI, Spleen-CT, and COVID-19 (CT) datasets. This visualization emphasizes the substantial presence of unlabeled data, approximately 60\% in each dataset, highlighting the relevance of our approach for practical clinical scenarios where pixel-wise annotated data for segmentation is scarce. The number of samples is presented in $\mathrm{x}$-axis}.
    \label{fig:falcon_data_splits}
\end{figure}

\subsection{Data Distribution and Preprocessing}
\label{subsec:falcon_data_dist_preprocessing}

\Cref{fig:falcon_data_splits} presents the sample distribution across these medical datasets used in our study, revealing a high proportion of unlabeled images: 60.74\%, 60.83\%, 60.37\%, and 59.80\% for CHAOS-CT, Spleen-CT, COVID-19 (CT), and Cardiac-MRI, respectively, within the training sets, underscoring the relevance of our approach for practical clinical scenarios where pixel-wise annotated data for segmentation is scarce.

\paragraph{Data preprocessing.} 
All medical imaging datasets consist of volumetric data, stored in either \texttt{DICOM} or \texttt{NIfTI} format. 2D slices are extracted from the 3D volumes and resized to $224 \times 224$ pixels. We use \texttt{pydicom} for handling \texttt{DICOM} files and \texttt{nibabel} for \texttt{NIfTI} files. Slices that are entirely black (i.e., with zero intensity across all pixels) or contain no anatomical content, such as those outside the region of interest (ROI) (e.g., top/bottom slices with only background), are removed during preprocessing.

\section{Experiments}
\label{sec:falcon_experiments}

\subsection{Implementation Details}
\label{subsec:implementation}

The Adam optimizer is used with a learning rate of 0.001 for both the training and fine-tuning phases. In accordance with the findings of \citet{Karimi2020}, the loss weight factors $\lambda_1$ and $\lambda_2$ are set to 0.9 and 0.1, respectively, while the parameter $a$ is set to 0.2. Experiments were conducted using an NVIDIA GeForce RTX 3070 GPU. Code is written in Python (v3.10) using PyTorch (v2.0) DL library.

%\enlargethispage{-\baselineskip}
\subsection{Evaluation Protocols}
\label{subsec:evaluation_protocols}

We employ two standard segmentation metrics: Dice Similarity Coefficient (DSC) and Hausdorff Distance (HD) to evaluate the segmentation performance of our framework. DSC is widely used in medical image segmentation to measure the overlap between the predicted and ground-truth regions. However, HD is of greater significance in scenarios like ours, where precise boundary delineation is the primary objective. 

\paragraph{Dice Similarity Coefficient.} Assuming two non-empty sets $\mathrm{U}$ and $\mathrm{V}$ contain the image pixels of the segmented area for the ground truth $\boldsymbol{y}_q$ and prediction map $\hat{\boldsymbol{y}}_q$, DSC is computed as:

\begin{equation}
    \mathrm{DSC}(\mathrm{U}, \mathrm{V}) = \frac{2\times \mathrm{U} \cap \mathrm{V}}{\mid \mathrm{U} \mid + \mid \mathrm{V} \mid}.
\end{equation}
Its range varies from 0 to 1, where 0 indicates no overlap between the sets, and 1 indicates a perfect overlap.

\paragraph{Hausdorff Distance.} In contrast to the DSC, assume the non-empty sets $\mathrm{U}$ and $\mathrm{V}$ containing boundary pixels for ground truth $\boldsymbol{y}_q$ and prediction map $\hat{\boldsymbol{y}}_q$, respectively. HD is then defined as the directed distance from $\mathrm{U}$ to $\mathrm{V}$, as follows:

\begin{equation}
    \mathrm{HD}_{\mathrm{U} \rightarrow \mathrm{V}} = \max_{\substack{u \in \mathrm{U}}} \min_{\substack{v \in \mathrm{V}}} \lVert u - v \rVert,
\end{equation}
where $u$ and $v$ are boundary pixels belonging to $\mathrm{U}$ and $\mathrm{V}$, respectively. We report the 95\textsuperscript{th} percentile of the HD, thereby discarding a small fraction of outliers, defined as:
\begin{equation}
    \mathrm{HD}^{95}_{\mathrm{U} \rightarrow \mathrm{V}} = \operatorname{percentile}_{95}\left( \left\{ \min_{v \in \mathrm{V}} \lVert u - v \rVert \,\middle|\, u \in \mathrm{U} \right\} \right).
\end{equation}

\section{Experimental Results} 
\label{sec:falcon_results}

This section presents the experimental results of FALCON. To validate our proposed approach and demonstrate the superior performance achieved by the FALCON architecture, we compare it against three other models: a \textbf{Baseline} trained with $\mathcal{L}_{\text{bce}}$, \textbf{Model A} trained with $\mathcal{L}_{dl}$, and \textbf{Model B} trained with $\mathcal{L}_{\text{hd}}$ to evaluate the impact of different segmentation loss functions empirically. All these models are evaluated across 10 FSL test tasks, and their average performance is reported in DSC and HD metrics. The test tasks are drawn from the test set, i.e., on patients unseen during meta fine-tuning within the target medical domain. Moreover, FALCON's performance is compared with SOTA models using the DSC metric. Most of these works did not report results using the HD metric, and their codebases are not publicly available for reproduction. Therefore, a direct comparison using the HD metric was infeasible.

\subsection{Quantitative Result}
\begin{table}[htbp] 
\caption[Quantitative results (DSC)]{Quantitative results in terms of \textbf{DSC} metric, comparing the performance of FALCON with the Baseline, Model A, and Model B, across four medical image segmentation problems: liver segmentation (CHAOS-CT), spleen segmentation (Spleen-CT), lung segmentation (COVID-19), and left atrium segmentation (Cardiac-MRI). The best results are highlighted in bold.} 
\label{tab:falcon_DSC}
\centering
\begin{tabularx}{0.9\textwidth}{@{}>{\raggedright\arraybackslash}X rrrr@{}}
\toprule
Model           & CHAOS-CT & Spleen-CT & COVID-19 & Cardiac-MRI \\
\midrule
Baseline ($\mathcal{L}_{bce}$) & 89.38 & 91.91 & 88.16 & 84.34 \\
Model A ($\mathcal{L}_{dl}$)   & 92.05 & 91.98 & 89.28 & 85.37 \\
Model B ($\mathcal{L}_{hd}$)~\citep{Karimi2020} & 88.27 & 91.03 & 89.62 & 81.98 \\ 
\midrule
FALCON (ours) & \textbf{93.86} & \textbf{93.34} & \textbf{90.74} & \textbf{85.97} \\
\bottomrule
\end{tabularx}
\end{table}

\begin{table}[htbp] 
\caption[Quantitative results (HD)]{Quantitative results in terms of \textbf{HD} metric, comparing the performance of FALCON with the Baseline, Model A, and Model B, across four medical image segmentation problems: liver segmentation (CHAOS-CT), spleen segmentation (Spleen-CT), lung segmentation (COVID-19), and left atrium segmentation (Cardiac-MRI). The best results are highlighted in bold.} 
\label{tab:falcon_HD}
\centering
\begin{tabularx}{0.9\textwidth}{@{}>{\raggedright\arraybackslash}X rrrr@{}}
\toprule
Model            & CHAOS-CT & Spleen-CT & COVID-19 & Cardiac-MRI \\
\midrule
Baseline ($\mathcal{L}_{bce}$) & 13.70 & 5.80 & 11.04 & 5.85 \\
Model-A ($\mathcal{L}_{dl}$)   & 13.17 & 4.41 & 8.22 & 5.60 \\
Model-B ($\mathcal{L}_{hd}$)~\citet{Karimi2020} & 13.01 & 3.90 & 6.67 & 5.06 \\ 
\midrule
FALCON (ours) & \textbf{10.78} & \textbf{3.32} & \textbf{4.91} & \textbf{4.30} \\
\bottomrule
\end{tabularx}
\end{table}

\Cref{tab:falcon_DSC,tab:falcon_HD} report the DSC and HD scores for FALCON, respectively, which compare the \textbf{Baseline}, \textbf{Model A}, and \textbf{Model B} across four medical image segmentation problems: liver segmentation (CHAOS-CT), spleen segmentation (Spleen-CT), lung segmentation (COVID-19), and left atrium segmentation (Cardiac-MRI). In our experiments, FALCON achieved the highest DSC and lowest HD values on all four datasets, indicating strong region-level overlap and precise boundary delineation. These results suggest that FALCON’s combination of architectural design, boundary-aware learning coupled with adversarial regularization, and training strategy effectively optimizes both anatomical area and segmentation boundaries. Model A consistently ranked second in DSC for the CHAOS-CT, Spleen-CT, and Cardiac-MRI datasets, while Model B slightly surpassed it on the COVID-19 dataset. For HD, where lower scores reflect better boundary accuracy, Model B outperformed Model A across all datasets but did not match FALCON. Overall, FALCON consistently outperforms the Baseline, Model A, and Model B, showing steady improvements in both DSC and HD, and culminating in strong performance across all metrics for these segmentation problems.

\begin{table}[!h]
\centering
\caption{SOTA comparison using DSC metric on the CHAOS-CT dataset.}
\label{tab:falcon_CHAOS_DSC}
\centering
\begin{tabularx}{0.6\textwidth}{@{} >{\raggedright\arraybackslash}X r@{}}
\toprule
Method & DSC$\uparrow$ \\
\midrule
PKDIA~\citep{kavur2021chaos} & \textcolor{blue}{97.79} \\ %$\pm$0.43 \\
Sli2Vol~\citep{yeung_sli2vol} & 91.00\\ %$\pm$2.9 \\
LE-UDA~\citep{zhao2022le} & 80.70 \\ %$\pm$9.8 \\
\midrule
FALCON (ours) & \textbf{93.86}\\ %$\pm$1.07} \\
\bottomrule
\end{tabularx}
\end{table}
%\vspace{1em}
\begin{table}[!h]
\caption{SOTA comparison (DSC) on the Spleen-CT dataset.}
\label{tab:falcon_Spleen_DSC}
\centering
\begin{tabularx}{0.6\textwidth}{@{} >{\raggedright\arraybackslash}X r@{}}
\toprule
Method & DSC$\uparrow$ \\
\midrule
C2FNAS-Panc~\citep{yuC2FNAS2020} & 96.60 \\
DiNTS~\citep{yufan2021_dints} & 96.98 \\
Swin-UNetR~\citep{tang_swin_unetr_2022} & 96.99 \\
Auto-nnU-Net~\citep{becktepe2025autonnunet} & 97.11 \\ %1.4 \\
Universal model~\citep{liu2024universal_model} & \textcolor{blue}{97.27} \\
\midrule
FALCON (ours) & \textbf{93.34} \\ %$\pm$1.52} \\
\bottomrule
\end{tabularx}
\end{table}

\begin{table}[!h]
\caption{SOTA comparison (DSC) on the COVID-19 (CT) dataset.}
\label{tab:falcon_COVID_DSC}
\centering
\begin{tabularx}{0.7\textwidth}{@{} >{\raggedright\arraybackslash}X r@{}}
\toprule
Method & DSC$\uparrow$ \\
\midrule
3D U-Net~\citep{ma2021toward} & 87.90 \\ %$\pm$9.3 \\
Cascaded U-Net~\citep{aswathy_2022} & 92.46 \\
SE-UNetR~\citep{pour_Beheshti} & 96.32 \\
SE-HQRSTNet~\citep{pour_Beheshti} & \textcolor{blue}{97.45} \\
\midrule
FALCON (ours) & \textbf{90.74} \\%$\pm$3.39} \\
\bottomrule
\end{tabularx}
\end{table}

\begin{table}[!h]
\centering
\caption{SOTA comparison (DSC) on the Cardiac-MRI dataset.}
\label{tab:falcon_CARDIAC_DSC}
\begin{tabularx}{0.6\textwidth}{@{} >{\raggedright\arraybackslash}X r@{}}
\toprule
Method & DSC$\uparrow$ \\
\midrule
MPUNet~\citep{perslev2019} & 89.00 \\ %± 0.09 \\
C2FNAS-Panc*~\citep{yuC2FNAS2020} & 92.49 \\
Swin-UNetR~\citep{tang_swin_unetr_2022} & \textcolor{blue}{94.80} \\
\midrule
FALCON (ours) & \textbf{85.97} \\%$\pm$2.68} \\
\bottomrule
\end{tabularx}
\end{table}

\paragraph{Comparison with state-of-the-art methods.} 
\Cref{tab:falcon_CHAOS_DSC,tab:falcon_Spleen_DSC,tab:falcon_COVID_DSC,tab:falcon_CARDIAC_DSC,} present the comparative DSC results of FALCON against state-of-the-art (SOTA) methods across these segmentation problems. SOTA results are derived from the original publications, and differences in experimental protocols should be anticipated when interpreting direct comparisons. In all tables, the best performance is highlighted in \textcolor{blue}{blue}, and FALCON’s results are shown in \textbf{bold}.

On \textbf{CHAOS-CT} (\Cref{tab:falcon_CHAOS_DSC}), FALCON achieved a DSC score of around 94\% without any augmentation, approximately 4\% lower than PKDIA, which was fully supervised and trained with extensive augmentation (scaling, rotation, shearing, thresholding). FALCON outperformed Sli2Vol, a self-supervised method, by approximately 4\% and exceeded LE-UDA, an unsupervised domain adaptation model, by about 14\%.

On \textbf{Spleen-CT} (\Cref{tab:falcon_Spleen_DSC}), FALCON’s performance was close ($\approx$3.5\%) to C2FNAS-Panc, DiNTS, and Auto-nnU-Net by approximately 3.5\%. These models leverage advanced neural architecture search (NAS). It was also close to Swin-UNetR, a transformer-based model by about 3.5\%. Furthermore, FALCON's performance was comparable to the Universal CLIP-driven model by approximately 4\%. 

On \textbf{COVID-19 (CT)} (\Cref{tab:falcon_COVID_DSC}), FALCON outperformed 3D U-Net and performed close  to the Cascaded 3D U-Net ($\approx$2\%). Computationally expensive transformer-based SE-variants, which incorporated extensive data augmentation, outperformed FALCON by about 7\%. Notably, FALCON achieved this accuracy using unlabeled data and without augmentation, preserving the structural realism critical in medical imaging.

On \textbf{Cardiac-MRI} (\Cref{tab:falcon_CARDIAC_DSC}), FALCON was close to MPUNet ($\approx$3\%). It trailed C2FNAS-Panc* ($\approx$ 6.5\%) and Swin-UNetR ($\approx$9\%), both computationally expensive models that employed data augmentation.

\begin{figure}[!ht]
    \centering
    \includegraphics[width=0.98\linewidth]{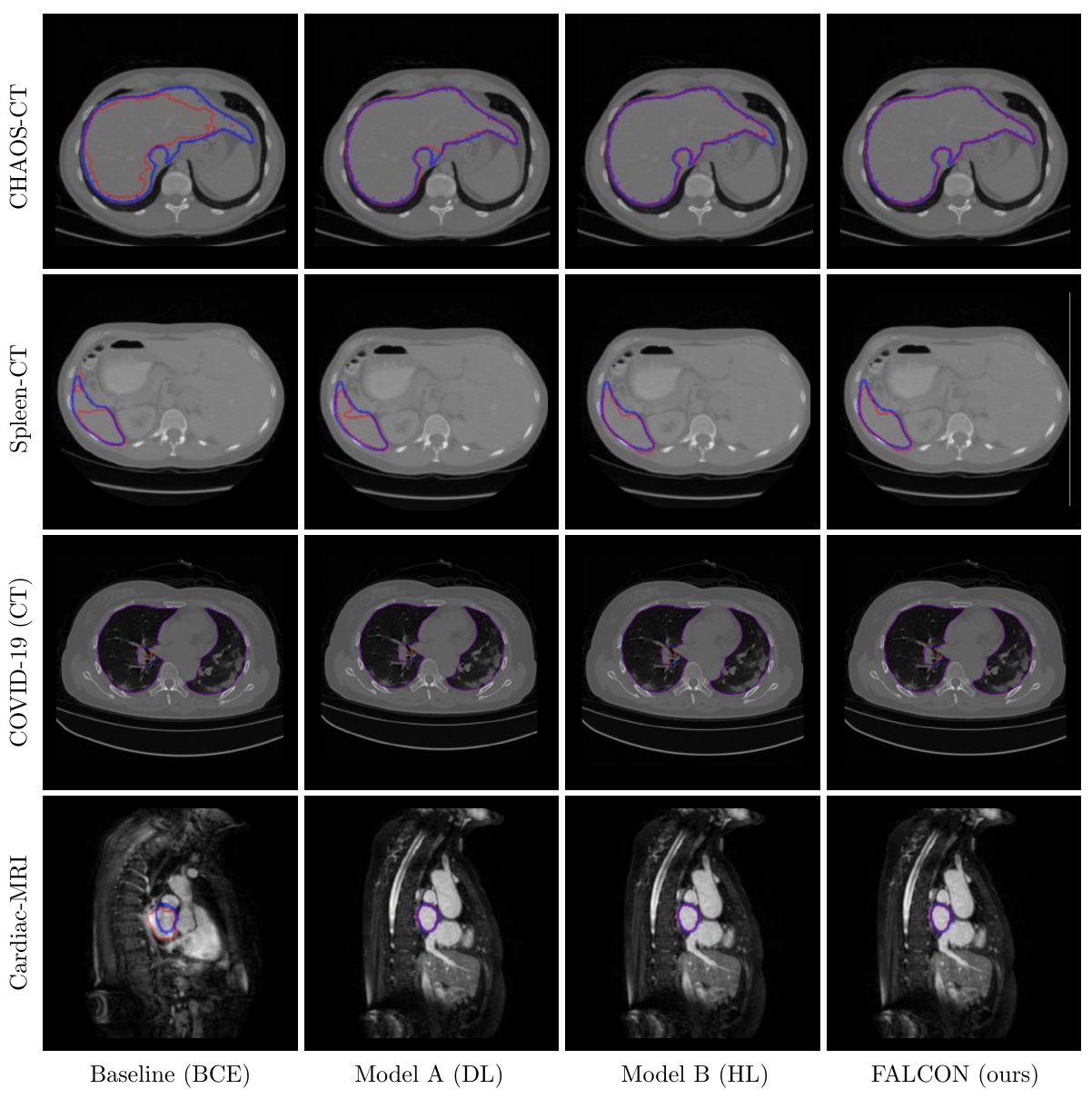}
    \caption[Qualitative results]{Qualitative results demonstrating the boundary delineation precision of our FALCON framework on the CHAOS-CT, Spleen-CT, COVID-19, and Cardiac-MRI datasets. Clinical ground truth annotations are shown in blue, while predicted segmentation maps are overlaid in red. Best viewed when zoomed in for clarity.}
    \label{fig:falcon_qualitative_results}
\end{figure}

\subsection{Qualitative Result}
\label{subsec:falcon_qualitative_results}

\Cref{fig:falcon_qualitative_results} presents qualitative results comparing FALCON's output (column 4) by comparing other models: Baseline (column 1), Model A (column 2) and Model B (column 3). This visualization also provides an understanding of the effect of the loss functions: Baseline, Model A and Model B employ BCE, Dice (DL), and Hausdorff loss (HL), respectively. Results are shown for the four segmentation problems involving CHAOS-CT, Spleen-CT, COVID-19, and Cardiac-MRI datasets in rows, with clinical ground truths marked in \textcolor{blue}{blue} and predictions in \textcolor{red}{red}. FALCON consistently produces smoother, anatomically coherent boundaries, particularly in regions with complex structures or sharp transitions, closely aligning predictions with ground truth. In contrast, the Baseline with BCE exhibits coarse and imprecise boundaries, while models trained with Dice or Hausdorff loss perform comparably across CHAOS-CT, COVID-19, and Cardiac-MRI datasets. Notably, Model B trained with HL preserves anatomical contours slightly better in the Spleen-CT dataset. The figure is best viewed at high magnification for detailed inspection of the boundary.

\subsection{Ablation studies}
\label{subsec:ablation}

\paragraph{Loss function analysis.} \Cref{tab:falcon_DSC,tab:falcon_HD} underscore the critical role of loss function selection in medical image segmentation, supported by \cref{fig:falcon_qualitative_results}, which illustrates that cross-entropy loss consistently underperforms across all medical test datasets. While Dice loss and Hausdorff loss achieve comparable overall performance, incorporating Hausdorff loss—especially when combined with adversarial learning within our framework—yields smoother, more precise boundaries, reflected in notably lower HD scores.

\paragraph{Relation module analysis.} To assess the contribution of the relation module—a core component of our FSL framework—we modified FALCON by removing it. Without this module, support features are no longer incorporated into the decoding process, and the model operates solely on query features, effectively reducing it to a standard U-Net-like semantic segmentation architecture. This change transitions the framework from a few-shot segmentation model to a conventional segmentation model, where the batch size equals the number of query images. \Cref{tab:falcon_RM_contribution} compares FALCON with and without the relation module. The results demonstrate that including the relation module substantially improves boundary precision, reducing HD scores by approximately 3.5 points on CHAOS-CT, 3 points on Spleen-CT, 2.5 points on COVID-19 (CT), and 5.5 points on the Cardiac-MRI dataset.

\begin{table}[ht]
    \caption[Relation module analysis]{Comparison of HD scores with and without (w/o) the relation module in the FALCON framework. The results highlight the contribution of the relation module to improving boundary precision by incorporating support features during decoding. Lower HD values indicate better boundary precision.}
    \label{tab:falcon_RM_contribution}
    \centering
    \begin{tabularx}{0.9\textwidth}{@{}>{\raggedright\arraybackslash}X rrrr@{}}
    \toprule
         & CHAOS-CT & Spleen-CT & COVID-19 & Cardiac-MRI \\
    \midrule
    FALCON (w/o)  & 14.26 & 6.38 & 7.45 & 9.75 \\
    FALCON (ours)  & 10.78 & 3.32 & 4.91 & 4.30 \\
    \bottomrule
    \end{tabularx}
\end{table}

Within our framework, the relation module acts as an implicit attention mechanism, enabling the model to learn a joint representation of support and query features. Although the support set is unlabeled, its feature representations carry rich structural cues, such as object boundaries, textures, and anatomical patterns, which are particularly informative in medical imaging. The module extracts and aligns these visual patterns to enhance the semantic understanding of the query input. Given our $1$-way $K$-shot binary segmentation setting, this alignment promotes feature-level consistency within a shared latent space, facilitating robust generalization across patients. Furthermore, the support-query joint encoding mitigates feature variability, allowing the model to calibrate query representations based on the more stable patient-specific characteristics present in the support set.

\begin{figure}[!htbp]
    \centering
    \includegraphics[width=0.99\linewidth]{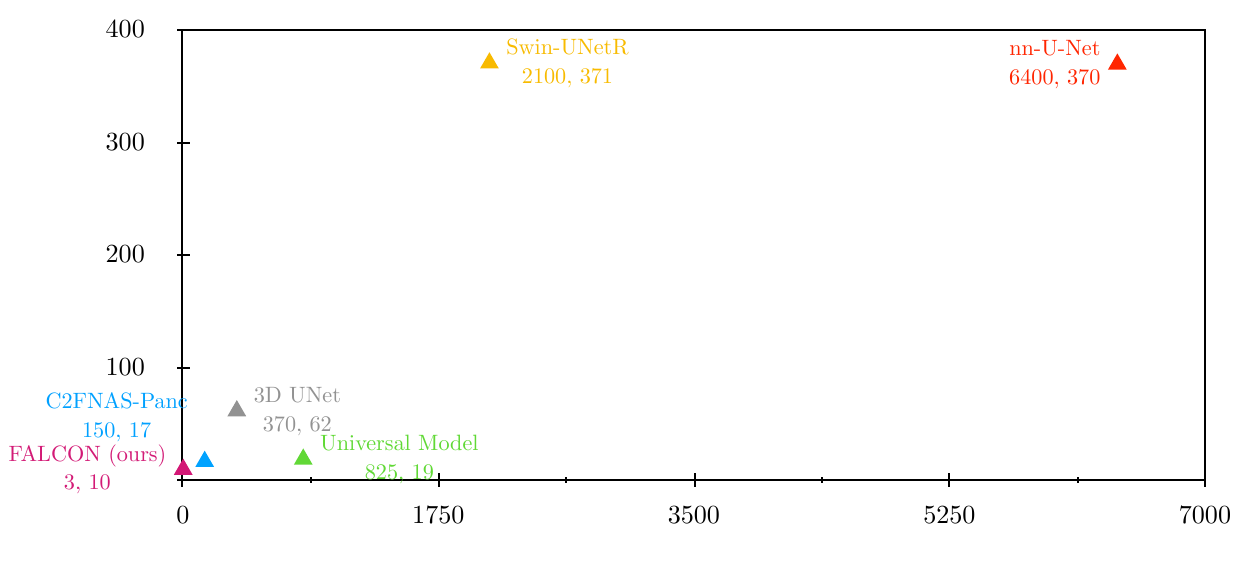}
    \caption[Computational complexity analysis]{Comparison of computational complexity across various SOTA models. The GFLOP is presented in $\mathrm{x}$-axis, and the number of parameters (in millions) is presented in $\mathrm{y}$-axis. Our proposed framework, FALCON, achieves competitive segmentation performance while maintaining significantly lower computational cost, with only 9.90 million parameters and 2.30 GFLOPs, compared to the high computational demands of models such as Swin-UNetR and nnU-Net.}
    \label{fig:falcon_compute_capability}
\end{figure}

\paragraph{Analysis of computational complexity and number of parameters.} We assess computational complexity in terms of GFLOPs, a standard measure of inference cost. FALCON requires only 2.30 GFLOPs with 9.90 million parameters, achieving competitive segmentation performance. In contrast, SOTA models such as C2FNAS-Panc (17M, 150 GFLOPs), 3D U-Net (19M, 825 GFLOPs), Universal model (62M, 370 GFLOPs), Swin-UNetR (371.94M, 2100 GFLOPs), and nnU-Net (370.74M, 6400 GFLOPs) demand orders of magnitude more resources. \Cref{fig:falcon_compute_capability} illustrates this comparison, highlighting that FALCON delivers high performance with dramatically lower computational cost.

\section{Discussion} 
\label{sec:falcon_discussion}

We hypothesized that a task-aware inference mechanism enables lightweight models to achieve performance comparable to state-of-the-art (SOTA) methods by leveraging the inherent structural consistency of unlabeled slices for volumetric segmentation of anatomical structures. Our results support this hypothesis: FALCON consistently achieves DSC scores within 3–5\% of (most) SOTA models across four segmentation tasks, despite using orders of magnitude fewer parameters and GFLOPs. This is particularly notable given that most SOTA methods are evaluated under ideal i.i.d. conditions—where training and test data share similar distributions—and often fail to maintain performance in real-world clinical workflows~\cite{google_med_test_MIT_news}, where imaging protocols, scanners, patient populations, and anatomical variations differ substantially

FALCON’s architecture is designed to address this challenge by formulating segmentation as a patient-specific FSL task, where each patient at inference is treated as a previously unseen `class.'  By integrating a relation module and training with a boundary-aware adversarial loss, it leverages unlabeled support slices from the target domain to calibrate query representations. This enables the model to align structural cues, such as organ boundaries and textures across slices within a patient, resulting in consistently lower HD scores and sharper anatomical boundaries. These improvements are not just numerical; they are clinically meaningful, as precise contours directly influence diagnostic reliability, longitudinal monitoring, and treatment planning. Moreover, the ability to adapt using only a handful of labeled samples underscores the framework’s practicality in annotation-scarce clinical environments.

Equally important is FALCON’s compact footprint (9.9M parameters, 2.3 GFLOPs), which makes it deployable on standard clinical hardware. Unlike transformer- or NAS-based models, FALCON does not rely on expensive computational infrastructure, facilitating broader adoption in resource-constrained environments. This efficiency also supports secure, local deployment, reducing dependence on cloud-based AI services and mitigating privacy concerns—a key barrier that frequently limits the clinical adoption of AI tools. Moreover, by leveraging natural image pretraining and adapting via unlabeled support for intra-patient context, FALCON demonstrates robustness to patient variability, while suggesting strong potential for cross-institutional and cross-modality applications.

In summary, while FALCON does not aim to replace large-scale SOTA architectures, it offers a practical, efficient, and clinically viable alternative that balances accuracy, adaptability, and deployability. Its core design principles—lightweight U-Net backbone, boundary-aware loss, and the ability to leverage unlabeled support slices point toward a promising direction for segmentation models intended for real-world clinical use under resource-constrained clinics. However, its current formulation is limited to 1-way (binary) segmentation, which, while common in clinical practice (e.g., organ or lesion delineation), restricts applicability in multi-organ or multi-class scenarios.

\section{Conclusion} 
\label{sec:falcon_conclusion}
In this study, we presented FALCON, an efficient CDFSL framework for medical image segmentation. Specifically engineered for resource-constrained clinical environments, FALCON operates under extreme label scarcity and leverages abundant unlabeled intra-patient data to achieve precise boundary delineation, enabled by a combination of relation-based contextual adaptation, adversarial regularization, and Hausdorff distance–aware optimization. By building on natural-image pretraining and adapting to the medical domain through boundary-aware adversarial fine-tuning, FALCON effectively bridges the domain gap without requiring large labeled medical datasets. With only 9.9 million parameters and 2.3 GFLOPs, FALCON enables privacy-preserving, on-device inference on standard clinical hardware, reducing reliance on cloud-based AI services and supporting the deployment of locally executable, patient-centric AI solutions.

\bibliography{ref}

\end{document}